%% file: main.tex
\documentclass[runningheads]{llncs}
\usepackage[T1]{fontenc}
\usepackage{blindtext}
\usepackage{graphicx}
\usepackage[colorlinks,urlcolor=blue]{hyperref}
\usepackage{color}
\usepackage{url}  

\usepackage{mathtools}
\usepackage{amsmath,amsfonts,amssymb}

\usepackage{tabularray}
\usepackage[misc]{ifsym}
\usepackage{booktabs}
\usepackage{upgreek}
\usepackage{enumitem}
\usepackage{bm}
\usepackage{censor}
\usepackage{textcomp}
\usepackage[english]{babel}
\usepackage{lastpage}
\usepackage[section]{placeins}
\usepackage{fancyvrb} 
\usepackage[dvipsnames,table]{xcolor}
\usepackage{fontawesome5}  

\usepackage{multirow}
\usepackage{makecell}
\usepackage{subfigure}
\usepackage{tabularx,ragged2e}
\usepackage{booktabs}

\definecolor{LightCyan}{rgb}{0.88,1,1}
\definecolor{lightskyblue}{RGB}{225, 235, 240}

\definecolor{Gray}{gray}{0.90}
\definecolor{white}{rgb}{1.0, 1.0, 1.0}

\definecolor{Lightgreen}{RGB}{218, 246, 230 }

\begin{document}
\title{SurvRNC: Learning Ordered Representations for Survival Prediction using Rank-N-Contrast}
\titlerunning{SurvRNC: Survival Prediction using Rank-N-Contrast}
%
%
\author{Numan Saeed$^*$\textsuperscript{\Letter} \and
Muhammad Ridzuan$^*$ \and
Fadillah Adamsyah Maani\and
Hussain Alasmawi \and
Karthik Nandakumar \and
Mohammad Yaqub}
\authorrunning{N. Saeed et al.}
%
\institute{Mohamed bin Zayed University of Artificial Intelligence, Abu Dhabi, UAE
\email{\{numan.saeed, muhammad.ridzuan, fadillah.maani, hussain.alasmawi, karthik.nandakumar, mohammad.yaqub\}@mbzuai.ac.ae}
}
%
%
\maketitle              
\def\thefootnote{*}\footnotetext{Equal contribution}

\begin{abstract}

Predicting the likelihood of survival is of paramount importance for individuals diagnosed with cancer as it provides invaluable information regarding prognosis at an early stage. This knowledge enables the formulation of effective treatment plans that lead to improved patient outcomes. In the past few years, deep learning models have provided a feasible solution for assessing medical images, electronic health records, and genomic data to estimate cancer risk scores. However, these models often fall short of their potential because they struggle to learn regression-aware feature representations. In this study, we propose Survival Rank-N-Contrast (SurvRNC) method, which introduces a loss function as a regularizer to obtain an ordered representation based on the survival times. This function can handle censored data and can be incorporated into any survival model to ensure that the learned representation is ordinal. The model was extensively evaluated on a HEad \& NeCK TumOR (HECKTOR) segmentation and the outcome-prediction task dataset. We demonstrate that using the SurvRNC method for training can achieve higher performance on different deep survival models. Additionally, it outperforms state-of-the-art methods by $3.6\%$ on the concordance index. The code is publicly available on \url{https://github.com/numanai/SurvRNC}.


\end{abstract}
\section{Introduction}
Survival prediction is fundamental and pervasive in medical care for various diseases, including cardiovascular, chronic respiratory, neurological diseases, and different types of cancer. With regard to cancer, the World Health Organization anticipated that in the year 2023, an estimated 20 million individuals would be diagnosed with cancer, resulting in approximately 10 million fatalities. This number is anticipated to increase to approximately 30 million new cases by 2040 \cite{pahoWorldCancer}. Survival prediction aids in treatment planning, patient staging, and monitoring, which results in better cancer patient care \cite{mariotto2014cancer}. In the era of precision medicine, data-driven techniques and deep learning models enable healthcare professionals to make precise predictions of future outcomes through survival models.

The main idea of survival modeling is the use of regression analysis, a statistical method that models the relationship between the continuous outcome of interest (i.e., risk score) and its covariates (i.e., patient data). However, survival prediction is a complex task influenced by various factors, including disease physiology, clinical demographics, and treatment plans. Furthermore, incomplete survival data owing to right-censored samples, in which the exact event occurrence time is missing, pose a challenge. This can occur when patients are lost to follow-up, or when an event is not observed within a limited follow-up time. Nevertheless, data from censored patients still holds valuable information, as these patients did not experience the event within a specific period and may have experienced it later. Therefore, utilizing both uncensored and censored patient data is crucial to maximize the use of limited available data in survival prediction models \cite{buckley1979linear}. 

Substantial research has focused on improving the performance of survival models and incorporating censored patient data. The Cox proportional hazard (CoxPH) regression model \cite{cox1972regression} is a widely used statistical technique for analyzing survival data with censored patients. However, it has limitations in that it assumes linear relationships between covariates, survival probability, and constant hazard ratios over time. An alternative to CoxPH is the multi-task logistic regression (MTLR) model \cite{yu2011learning}, which is more flexible and can model non-proportional hazards and the complex relationships between covariates and survival outcomes. Additionally, survival support vector machines \cite{polsterl2015fast} and random survival forests \cite{ishwaran2008random} are two other methods that can be used to model nonlinear, complex relationships between covariates and future outcomes.

In the oncology context, survival models must distill complex high-dimensional multimodal data, including imaging (e.g., computed tomography (CT) or positron emission tomography (PET) scans), clinical (e.g., electronic health records (EHR)), and molecular data (e.g., genomics), into actionable insights. Therefore, it is necessary to employ deep learning to extract representations across the different input views in a survival prediction framework. Deep survival models use convolutional neural networks (CNNs) or vision transformers (ViTs) to extract features from images and fuse them with EHR before feeding them to a survival prediction model. Various studies have used deep learning-based survival prediction models for the prognosis of oral \cite{kim2019deep}, ovarian \cite{zheng2022preoperative}, glioma \cite{choi2022estimating}, gastric \cite{hu2022deep}, breast \cite{ng2023deep}, and prostate \cite{elmarakeby2021biologically} cancers. Recently, the prognosis of lung cancer was investigated through the use of image-based features extracted via deep learning classification or segmentation models \cite{wang2019artificial}. In another context, breast cancer prognosis has been attempted in several studies, where deep-learning-aided feature extraction is used, leveraging genomic data and pathological images \cite{wang2021gpdbn}, RNA-seq data \cite{huang2020deep}, and mammograms \cite{arefan2020deep}. Recently, there has been a lot of interest in modeling the future outcomes of patients with Head and Neck (H\&N) cancer, which is one of the most common cancers worldwide. In \cite{meng2023merging}, an X-shape hybrid transformer network (XSurv) was proposed, which consists of merging an encoder for multimodal data fusion and a decoder for survival prediction. In another work \cite{saeed2022tmss}, a transformer-based implementation called TMSS provides an end-to-end solution that takes CT and PET scans, as well as EHR data, and not only predicts the survival probability but also segments the H\&N tumor. 

However, the existing deep learning-based survival prediction methods solely target an end-to-end solution for future outcome predictions without explicitly focusing on learning regression-aware feature representations. We argue that by not making this consideration (i.e., constraining the learned representation properly), deep learning methods underperform as they remain oblivious to the ordinal nature of the problem. In recent work \cite{zha2024rank}, an extension of the contrastive loss function called Rank-N-Contrast (RNC) was introduced to learn a regression-aware representation by contrasting samples against each other based on their rankings in the target space. However, this cannot be applied directly to a survival prediction problem due to the presence of censored patients in the dataset. 

In this study, we propose the \textbf{Surv}ival \textbf{R}ank-\textbf{N}-\textbf{C}ontrast (SurvRNC) method, a novel approach that incorporates a unique loss function to learn ordinal feature representations based on survival times. This function is capable of handling censored data and can be incorporated into any survival model to ensure the learned representation is ordinal. Our research demonstrates that SurvRNC enhances the performance of various deep learning models, including state-of-the-art DeepMTLR \cite{deepmtlr} and DeepHit \cite{deephit}, for complex tasks such as predicting survival rates in head and neck cancer cases using multimodal datasets.

\section{Methodology}

\subsection{Problem Statement}

We aim to develop a deep neural network that utilizes multimodal data, specifically CT/PET scans and EHR, to predict patient survival through learning an ordered feature representation. The dataset for this task, denoted as $D = [\mathcal{P}_1, \mathcal{P}_2, ..., \mathcal{P}_N]$, consists of $N$ patients, where a patient $\mathcal{P}_j = (X^{j}, e^{j}, T^{j})$, comprises patient features $X^{j}$ , event indicator $e^{j}$ (with $e=0$ and $e=1$ indicating censored and uncensored patients, respectively), and time-to-event $T^{j}$. The objective is to train the network to predict the survival probability of a patient beyond a certain time $t$, denoted as $S(t|X) = P\left(T > t, e \mid X\right)$, which corresponds to the patient's \textit{survival function}. 

\begin{figure}[!tp]
    \centering
    \includegraphics[width=\linewidth]{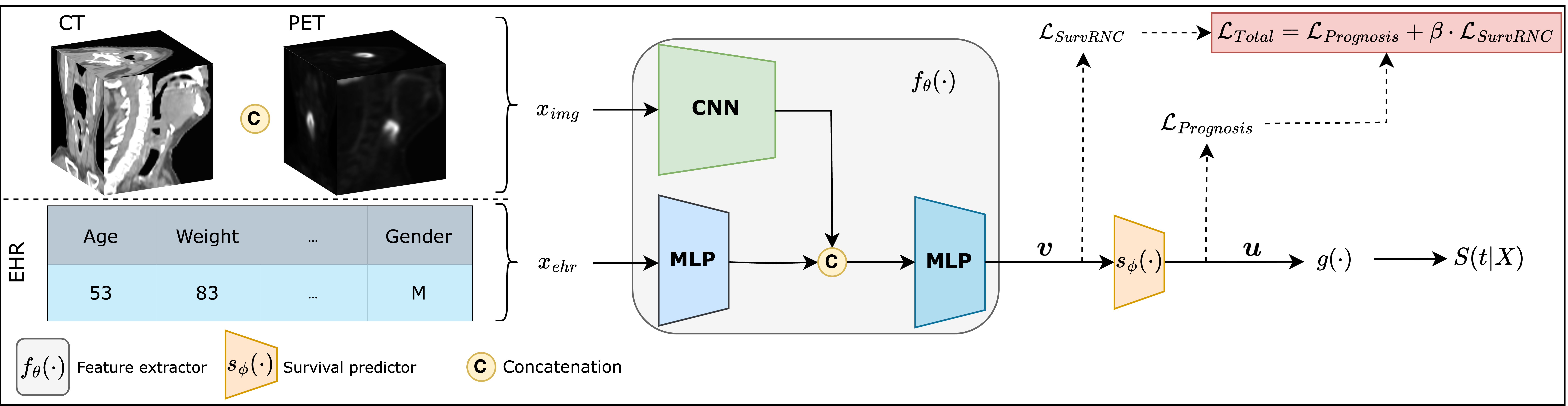}
    \caption{An illustration of the deep neural network architecture trained using its native loss $\mathcal{L}_{Prognosis}$ and $\mathcal{L}_{SurvRNC}$.}
    \label{fig:arch}
\end{figure}

The deep neural network shown in Figure \ref{fig:arch} consists of two components: a feature encoder, denoted by $f_{\theta}(X) \rightarrow \boldsymbol{v} \in \mathbb{R}^{d_{emb}}$, and a survival predictor $s_\phi(\boldsymbol{v}) \rightarrow \boldsymbol{u} \in \mathbb{R}^{d_t}$. The feature encoder $f_\theta$ takes as input $\textit{X} = [x_{img},x_{ehr}]$ and outputs representations $\boldsymbol{v}$ in the embedding space. These feature representations $\boldsymbol{v}$ are then passed to the survival predictor $s_\phi$ to predict $\boldsymbol{u}$, which can be a survival or risk score. For a patient, the survival function, $S(t|X)$, can be estimated using the output $\boldsymbol{u}$ and function $g(\cdot)$. To encourage the model to learn regression-aware feature representations $\boldsymbol{v}$, they must be appropriately ordered based on the censoring information $e$ and time-to-event $T$ labels.

\begin{figure}[!htbp]
    \centering
    \includegraphics[width=\textwidth]{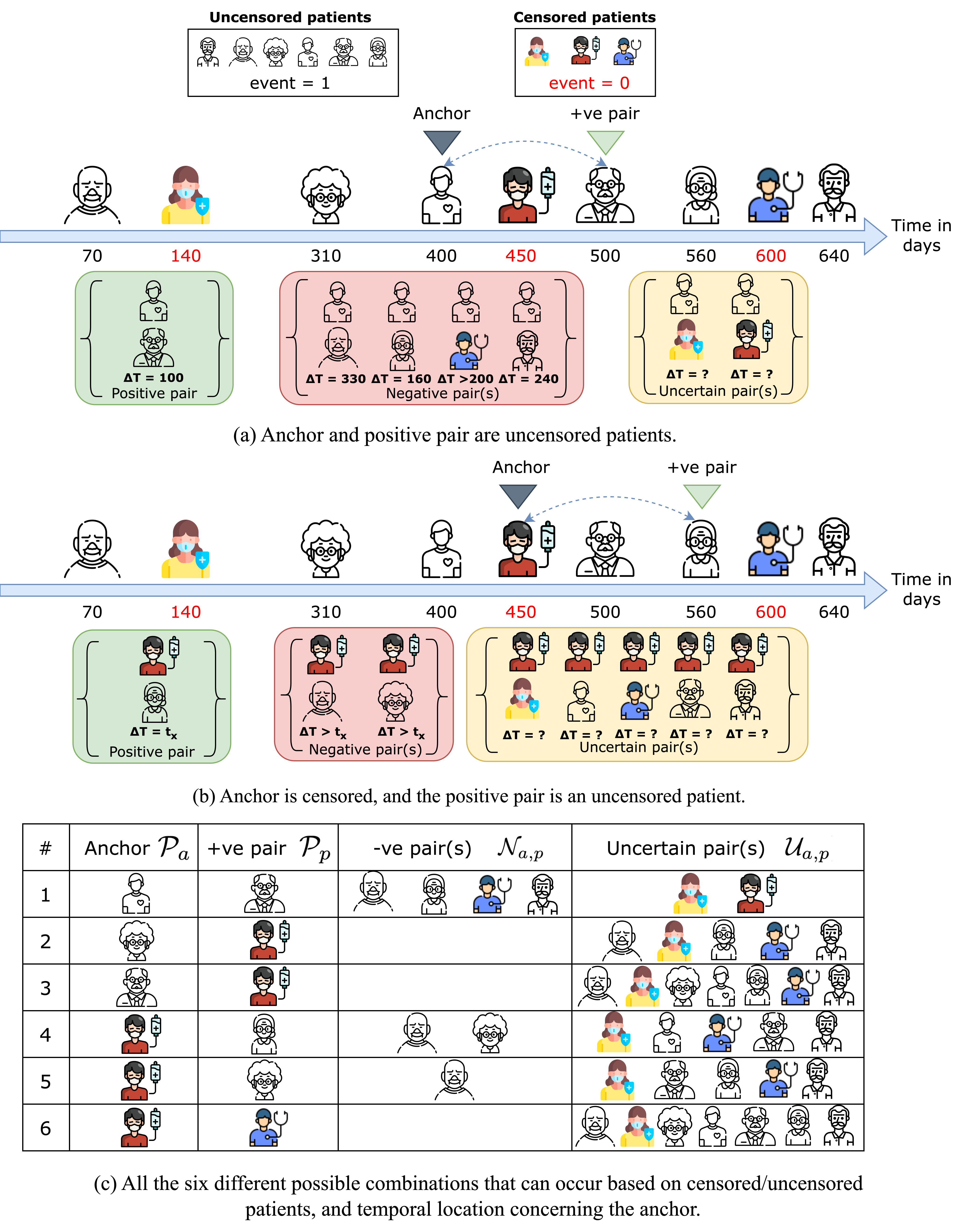}
    \caption{Overview of the proposed $\mathcal{L}_{SurvRNC}$ loss function for learning ordinal representations. In a randomly weighted sampled batch of $M$ patients, the loss function ranks them with respect to their time-to-event differences with the anchor. Contrasting the anchor patient with a positive pair patient enforces the similarity in the embedding space to be higher than the negative pair(s) of patients with a larger time-to-event difference than the positive pair. The uncertain patient pair(s), whose real-time difference with the anchor patient is unknown, are given less weight. (a) shows an example of an uncensored anchor and positive-pair patient. (b) shows an example of a censored anchor with an uncensored, positive pair. (c) provides all different combinations that can occur between a batch of patients.}
    \label{fig:lsurv_fig}
\end{figure}

\subsection{Survival Rank-N-Contrast Loss}
In order to learn an ordered feature representation, we propose \textbf{Surv} \textbf{R}ank-\textbf{N}-\textbf{C}ontrast, \textbf{$\mathcal{L}_{SurvRNC}$}, a loss function that ranks patients' representations in the embedding space based on their true time-to-event differences. Given an anchor patient, $\mathcal{P}_a$, we estimate the exponential increase in the likelihood of any other positive pair patient, $\mathcal{P}_p$, based on their similarity in the latent representation space. The time-to-event difference between $\mathcal{P}_a$ and $\mathcal{P}_p$ is denoted by $\Delta T_{a,p} \triangleq |T^a - T^p| $. The set of patients $k$ in the sampled batch that are higher in ranks than $\mathcal{P}_p$ in terms of their time-to-event differences from the anchor patient $\mathcal{P}_a$ is denoted by, $\mathcal{S}_{a,p} \triangleq$ \{$k: k \in [1,N] \setminus\{a\}, \Delta T_{a,k} \geq \Delta T_{a,p}$\}. Hence, the normalized likelihood $\ell$ is given by

\begin{equation}
\label{theta_like}
     \ell_{a,p} = \frac{\exp \left(\operatorname{sim}\left(\boldsymbol{v}_a, \boldsymbol{v}_p\right) / \tau\right)}{\sum_{k \in \mathcal{S}_{a, p}} \exp \left(\operatorname{sim}\left(\boldsymbol{v}_a, \boldsymbol{v}_k\right) / \tau\right)}
\end{equation}
where $sim(\cdot,\cdot)$ and $\tau$ denote the similarity measure $(L_2)$ and temperature parameter, respectively. Patients whose $\Delta T < \Delta T_{a,p}$ are disregarded in the equation. Maximizing $\ell$ in Equation \ref{theta_like} brings the feature representations $\boldsymbol{v}_a$ and $\boldsymbol{v}_p$ closer compared to the other representations $\boldsymbol{v}_k$ in the set $\mathcal{S}_{a,p}$. Typically, all patients in the set $\mathcal{S}_{a,p}$ are considered negative pair(s), but the presence of censoring in survival prediction datasets presents some unique circumstances due to uncertain $\Delta T$. Consequently, the set $\mathcal{S}_{a,p}$ now includes both the negative pair $(\mathcal{N}_{a,p})$ and uncertain pair $(\mathcal{U}_{a,p})$ sets, i.e. $\mathcal{S}_{a,p} = \{\mathcal{N}_{a,p} \cup \mathcal{U}_{a,p}\}$ 

Consider the example in Figure \ref{fig:lsurv_fig}, in which we have nine patients in a batch, three of whom are censored. Figure \ref{fig:lsurv_fig}(a) shows a scenario where the anchor patient $\mathcal{P}_a$ and the positive pair $\mathcal{P}_p$ are both uncensored with a time difference $\Delta T_{a,p}=100$. In this case, four pairs of patients with $\Delta T_{a,k} > 100$ will become part of the negative pair set $\mathcal{N}_{a,p}$, and two pairs of patients will become part of the uncertain pair set $\mathcal{U}_{a,p}$. Note that for the latter, their real time-to-events are unknown due to censoring. Thus, they can have $\Delta T \le 100$ or $\Delta T > 100$, i.e., to be disregarded from the equation or part of the negative pairs. Similarly, Figure \ref{fig:lsurv_fig}(b) shows a scenario where $\mathcal{P}_a$ is censored and $\mathcal{P}_p$ is uncensored, and the corresponding pairs that belong to $\mathcal{N}_{a,p}$ and $\mathcal{U}_{a,p}$. In total, we have six possible combinations of $(\mathcal{P}_a,\mathcal{P}_p)$ in any batch, as shown in Figure \ref{fig:lsurv_fig}(c), based on the censoring status of $\mathcal{P}_a$ and $\mathcal{P}_p$ and their temporal positions relative to one another.

 Consequently, when working with a randomly weighted sample batch $M$ of input, event, and target pairs, $\{(X^{j}, e^{j}, T^{j})\}_{j\in [M]}$, we first apply data augmentations to produce a two-view batch $\{(\tilde{X}^{j}, e^{j}, T^{j})\}_{j\in [2M]}$. Due to the presence of uncertain pairs in $\mathcal{U}_{a,p}$, $\hat{\ell}$ can be formulated as
\begin{equation}
    \hat{\ell}_{a,p}= \frac{\exp \left(\operatorname{sim}\left(\boldsymbol{v}_a, \boldsymbol{v}_p\right) / \tau\right)}{\sum_{k \in \mathcal{N}_{a, p}} \exp \left(\operatorname{sim}\left(\boldsymbol{v}_a, \boldsymbol{v}_k\right) / \tau\right) + \lambda {\sum_{k \in \mathcal{U}_{a, p}} \exp \left(\operatorname{sim}\left(\boldsymbol{v}_a, \boldsymbol{v}_k\right) / \tau\right)}} 
\end{equation}
 The loss function $\mathcal{L}_{SurvRNC}$ that is used to learn an ordered representation in the embedding space is given by
\begin{equation}
\label{lsurv}
\mathcal{L}_{\mathrm{SurvRNC}}=\frac{1}{2 M} \sum_{a=1}^{2 M} \frac{1}{2 M-1} \sum_{p=1, p \neq a}^{2 M}-\log \mathbb{\hat{\ell}}_{a,p}\\
\end{equation}

Intuitively, by minimizing the loss function, it encourages the similarity between an anchor patient $\mathcal{P}_a$ and a positive pair patient $\mathcal{P}_p$ to be larger than the similarity between anchor patient $\mathcal{P}_a$ and any other patient $\mathcal{P}_k$ in the set $\mathcal{S}_{a,p}$. 
The parameter $\lambda \in [0,1]$ in the $\mathcal{L}_{SurvRNC}$ allows us to control the weightage given to the uncertain pairs in set $\mathcal{U}_{a,p}$. By setting $\lambda = 0$, the pairs in $\mathcal{U}_{a,p}$ are disregarded, and if $\lambda=1$, then they are given equal weightage as the pairs in the negative pair(s) set $\mathcal{N}_{a,p}$. The total loss $\mathcal{L}_{Total}$ used to optimize the deep neural network-based survival model will consist of the native prognosis loss function used to optimize a specific survival model and the proposed $\mathcal{L}_{SurvRNC}$. The hyperparameter $\beta$, provides weightage to $\mathcal{L}_{SurvRNC}$.

\begin{equation}
    \mathcal{L}_{Total} = \mathcal{L}_{Prognosis} + \beta * \mathcal{L}_{SurvRNC}
\end{equation}

\section{Experimental Setup}

\subsection{Configurations, Implementation and Baselines}
We experiment with two different deep neural network-based survival models, i.e., DeepMTLR \cite{deepmtlr}, and DeepHit \cite{deephit} and investigate the impact of incorporating $\mathcal{L}_{SurvRNC}$ on the prediction performance. We also compare the proposed method with several top-performing survival prediction techniques for H\&N cancer, including the Cox-PH \cite{cox1972regression} method and individual coefficient approximation for risk estimation (ICARE) \cite{rebaud2022simplicity}. Additionally, the proposed method was contrasted with other advanced models, such as DeepMTLR-CoxPH \cite{saeed2021ensemble}, DeepMTS \cite{meng2022deepmts}, TMSS \cite{saeed2022tmss}, and XSurv \cite{meng2023merging}.

A uniform framework is defined to present a standardized testing environment for all networks, ensuring equality and impartiality with respect to patch dimensions, configuration, augmentations, training, and assessment. We use PyTorch \cite{paszke2019pytorch} for implementing the proposed framework. The models are trained for 50 epochs, with a batch size of 32, learning rate of $1\times10^{-4}$, and weight decay of $1\times10^{-5}$. A simple deep neural network has been designed, as shown in Figure \ref{fig:arch}. The feature extractor $f_\theta(\cdot)$ consists of a CNN block that extracts representations from CT and PET scans in a latent space and an MLP block that obtains representations of EHR data in the same latent space. These latent representations are then combined and fed into a survival prediction model  $s_\phi(\cdot)$ for making survival probability predictions. $\mathcal{L}_{SurvRNC}$ is applied in the embedding space after $f_\theta(\cdot)$. The architecture remains the same for all our experiments, except for the survival predictor $s_\phi(\cdot)$ at the head, which can be DeepMTLR \cite{deepmtlr} or DeepHit \cite{deephit}.

\input{tables/main_result}

\subsection{Dataset}
The dataset used is called HECKTOR (Head \& neCK TumOR segmentation and outcome prediction) \cite{andrearczyk2021overview}, a multi-modal and multi-center head and neck cancer patient data. The data consists of CT and PET scans with their segmentation masks and EHR for 488 patients. This dataset was collected from seven centers and contains Recurrence-Free Survival (RFS) data, including time-to-event and censoring status. Approximately $70\%$ of the patients in the dataset are censored. The clinical indicators distributions are provided in Appendix Table \ref{table:clinical_characteristics}, and the reprocessing steps and augmentation details are provided in Appendix Table \ref{my-label}.

\section{Results and Discussion}

Table \ref{tbl:hecktor_main} compares the performance of different deep survival-based models on a 5-fold cross-validation when they are trained using only their respective native losses and when they are trained including the proposed loss term, $\mathcal{L}_{SurvRNC}$. The evaluation metric is the concordance index (CI) and AUC at three points in the future, specifically 25\%, 50\%, and 75\% of the maximum duration. It can be observed that including the proposed $\mathcal{L}_{SurvRNC}$ boosts the performance in all the considered models. This is because the proposed loss is ordering the latent representation, which in turn helps achieve higher performance in survival predictions.

In recent work \cite{meng2023merging}, the authors used the same dataset and evaluated the performance of different state-of-the-art survival prediction models. In their work, patients from two centers (CHUM and CHUV) were used for testing, and other centers' data were used for training, which split the data into 386/102 patients in training/testing sets. Table \ref{tbl:soa_models} compares the performance of the DeepHit model trained using its native loss and $\mathcal{L}_{SurvRNC}$ on this train-test split. It is shown that the proposed method gives the highest CI on this test set compared to all the rest of the methods, which even includes complex architectures such as TMSS \cite{saeed2022tmss}, DeepMTS \cite{meng2022deepmts}, and XSurv \cite{meng2023merging}. One point to highlight here is that all these complex models are trained using both the segmentation masks and time-to-event labels. In contrast, the proposed method achieves better performance with a simple model architecture and approach without utilizing segmentation masks, showing the effectiveness of regularizing the latent representations by ordering them.

Despite a good performance on this particular train-test split, we believe that this is an easy test set as it does not adequately capture the complexity of the overall dataset, as can be observed by comparing Table \ref{tbl:soa_models} results with the 5-fold cross-validation results in Table \ref{tbl:hecktor_main}. Hence, we repeated our evaluation (i.e., involving the native loss alone and including $\mathcal{L}_{SurvRNC}$) on the private test set by submitting the predictions on their online portal. The test set consists of 339 patients from three different centers (out of which two centers are not part of the training set), which makes it a difficult test set. The CI score on the test set for the proposed method is 0.66, higher than any other method using only time-to-event labels for training. Notably, this result is on par with the best model from the HECKTOR leaderboard (which is trained using both segmentation mask and time-to-event labels, unlike ours, which uses time-to-event labels only).

We believe that these experiments and results show the effectiveness of the proposed methodology, where $\mathcal{L}_{SurvRNC}$ loss function is used to order the latent representations based on the time-to-event and censored label information. The ablation study Table \ref{tbl:hecktor_ablation} in Appendix shows that the weightage parameter $\lambda = 0.5$ gives the best concordance index, which is in line with the assumption that the patients in uncertain pair set ($\mathcal{U}_{a,p}$) has an equal probability of being in negative pair set ($\mathcal{N}_{a,p}$) or disregarded in the $\mathcal{L}_{SurvRNC}$ calculations. In addition, we present further evidence of continuity in the learned representations using a larger dataset with less proportion of censored patients (SUPPORT \cite{kim2019deep}) in Figure \ref{fig:support} of Appendix.

\section{Conclusion}
This work introduces a novel loss function called SurvRNC to order the latent representation according to time-to-event target labels for a survival prediction task. Exhaustive experimental results demonstrate that the proposed function surpasses existing state-of-the-art methods on the extensively benchmarked HECKTOR dataset for survival prediction. Future research may explore the potential of the SurvRNC method by incorporating it into end-to-end segmentation and prognosis models. Furthermore, the proposed approach could be applied to similar tasks with alternative datasets to evaluate its generalizability.


%
%
%
%

\bibliographystyle{splncs04}
\bibliography{ref}

\newpage
\section{Appendix}

\input{tables/appendix_A}

\end{document}

%% file: tables/main_result.tex
\begin{table*}[!t]
\centering
\begin{minipage}{0.45\textwidth}
\centering
\caption{Performance of different deep survival models on 5-fold cross-validation of hecktor dataset. }
\label{tbl:hecktor_main}
\setlength{\tabcolsep}{4pt}
\scalebox{0.8}{
\begin{tabular}{l | c | c  c  c} 
\toprule
 \rowcolor{Gray}
 {} & \textbf{CI $\uparrow$} & \multicolumn{3}{c}{\textbf{AUC $\uparrow$}} \\
 \rowcolor{Gray}
 \multirow{-2}{*}{\makecell[l]{ \text{Metrics $\rightarrow$} \\ 
 \text{Models $\downarrow$} }} & {} & 25\% & 50\% & 75\%  \\
 \midrule
 \textbf{DeepMTLR \cite{deepmtlr}} \\
 $\mathcal{L}_{MTLR}$ & 0.634\scriptsize±0.03     & 0.680 & 0.684 & 0.661 \\
 \rowcolor{lightskyblue}  $\mathcal{L}_{MTLR} + \mathcal{L}_{SurvRNC}$ 
 & 0.701\scriptsize±0.04         & 0.727 & 0.713 &  0.707  \\
 \midrule
 \textbf{DeepHit \cite{deephit}} \\
  $\mathcal{L}_{Hit}$ & 0.661\scriptsize±0.07     & 0.683 & 0.641 &  0.563  \\
 \rowcolor{lightskyblue} $\mathcal{L}_{Hit} + \mathcal{L}_{SurvRNC}$ 
  & 0.723\scriptsize±0.03         & 0.715 & 0.709 &  0.756   \\
 \bottomrule
\end{tabular}}
\end{minipage}
\hspace{3em}
\begin{minipage}{0.45\textwidth}
\centering
\caption{\small{Comparison with state-of-the-art survival prediction methods in leave two center out evaluation.}}
\label{tbl:soa_models}
\setlength{\tabcolsep}{4pt}
\scalebox{0.8}{
\begin{tabular}{ l c | c } 
\toprule
\rowcolor{Gray}
 \multicolumn{2}{c|}{Methods} & $\mathrm{\textbf{CI}}\hspace{-0.3em}\uparrow$ \\
\midrule
 CoxPH \cite{cox1972regression}    & Radiomics &  0.745\scriptsize±0.02 \\ 
 ICARE \cite{rebaud2022simplicity}   & Radiomics &  0.765\scriptsize±0.02 \\ 
Ensemble \cite{saeed2021ensemble}
& CNN &  0.748\scriptsize±0.03 \\ 
 TMSS \cite{saeed2022tmss}    & ViT+CNN &  0.761\scriptsize±0.03 \\
DeepMTS \cite{meng2022deepmts}  & CNN  &  0.757\scriptsize±0.02 \\ 
 XSurv \cite{meng2023merging} & Hybrid  & 0.782\scriptsize±0.02 \\
Radio-XSurv \cite{meng2023merging} & Hybrid+Rad.  & 0.798\scriptsize±0.02\\
 \hline
\rowcolor{lightskyblue}
SurvRNC & CNN  & 0.827\scriptsize±0.032\\
 \bottomrule
\end{tabular}}
\end{minipage}

\end{table*}

%% file: tables/appendix_A.tex
\begin{table} [!h]
\caption{Clinical characteristics of patients in the dataset.}
\label{table:clinical_characteristics}
\centering
\begin{tabular}{lcc}
\toprule
{\textbf{Characteristics}} & {\textbf{Dataset (n = 488)}} \\
\midrule
\textbf{RFS } \\
 Uncensored&  96 (19.7) \\
 Censored &   392 (80.3) \\
\midrule
\textbf{Age (year)} &    \\
\textbf{Weight (kg)} &   \\
\midrule
\textbf{Gender} &  & \\
Male & 402 (82.4) \\
Female & 86 (17.6) \\
\midrule
\textbf{Alcohol consumption} &  & \\
Yes & 95 (19.5) \\
No & 59 (12.1) \\ 
Unknown  & 334 (68.4) \\
\midrule
\textbf{Tobacco consumption} &  & \\
 Yes  & 85 (17.4) \\
 No & 105 (21.5) \\
 Unknown & 298 (61.1) \\
\midrule
\textbf{HPV status}&  & \\
 Positive & 274 (56.1) \\
 Negative & 43 (8.8) \\
 Unknown & 171 (35.1) \\
\midrule
\textbf{Performance status} &  & \\
 0 & 86 (17.7) \\
 1 & 114 (23.3) \\
 2 & 11 (2.3) \\
 3 & 3 (0.6) \\
 4 & 1 (0.2) \\
 Unknown & 273 (55.9) \\
\midrule
\textbf{Surgery} &  & \\
 Yes & 50 (10.3) \\
 No & 248   (50.8) \\
 Unknown & 190 (38.9) \\
\midrule
\textbf{Chemotherapy} &  & \\
 Yes  & 422 (86.5) \\
 No & 66 (13.5) \\
\bottomrule
\end{tabular}
\end{table}

\begin{table}[!h]
\centering
\caption{Preprocessing and augmentation details.}
\label{my-label}
\begin{tabularx}{\textwidth}{c *3{>{\Centering}X}}
\toprule
\bfseries Augmentations       & \bfseries Axis & \bfseries Probability & \bfseries Size       \\ \midrule
Orientation         & PLS  & -           & -          \\
CT/PET Concatenation & 1    & -           & -          \\
Normalization       & -    & -           & -          \\
Random crop         & -    & 0.5         & 96 x 96 x 96 \\
Random flip         & x, y, z & 0.1       & -          \\
Rotate by 90 (up to 3x) & x, y & 0.1       & -          \\ \bottomrule
\end{tabularx}
\end{table}

\begin{table}[!h]
\centering
\caption{Ablation to study the effect of the parameter $\lambda$ values.}
\label{tbl:hecktor_ablation}
\setlength{\tabcolsep}{4pt}
\scalebox{1.0}{
\begin{tabularx}{\textwidth}{l | *4{>{\Centering}X}}
\toprule
 \rowcolor{Gray}
 {} & \multicolumn{4}{c}{\textbf{$\lambda$}} \\
 \rowcolor{Gray}
 \multirow{-2}{*}{\makecell[l]{ \text{} \\ 
 \text{Models} }} & 0.3 & 0.5 & 0.7 & 1.0  \\
 \midrule
  DeepMTLR \cite{deepmtlr}\\ $\mathcal{L}_{MTLR} + \mathcal{L}_{SurvRNC}$ 
  & 0.6870 & \cellcolor{lightskyblue}0.7009 & 0.6690 & 0.6742   \\
 \midrule
 DeepHit \cite{deephit} \\ $\mathcal{L}_{Hit} + \mathcal{L}_{SurvRNC}$ 
  & 0.6802 & \cellcolor{lightskyblue} 0.7233 & 0.6950 & 0.6808   \\
 \bottomrule
\end{tabularx}}
\end{table}

\begin{figure}[!h]
    \centering
    \includegraphics[width=\linewidth]{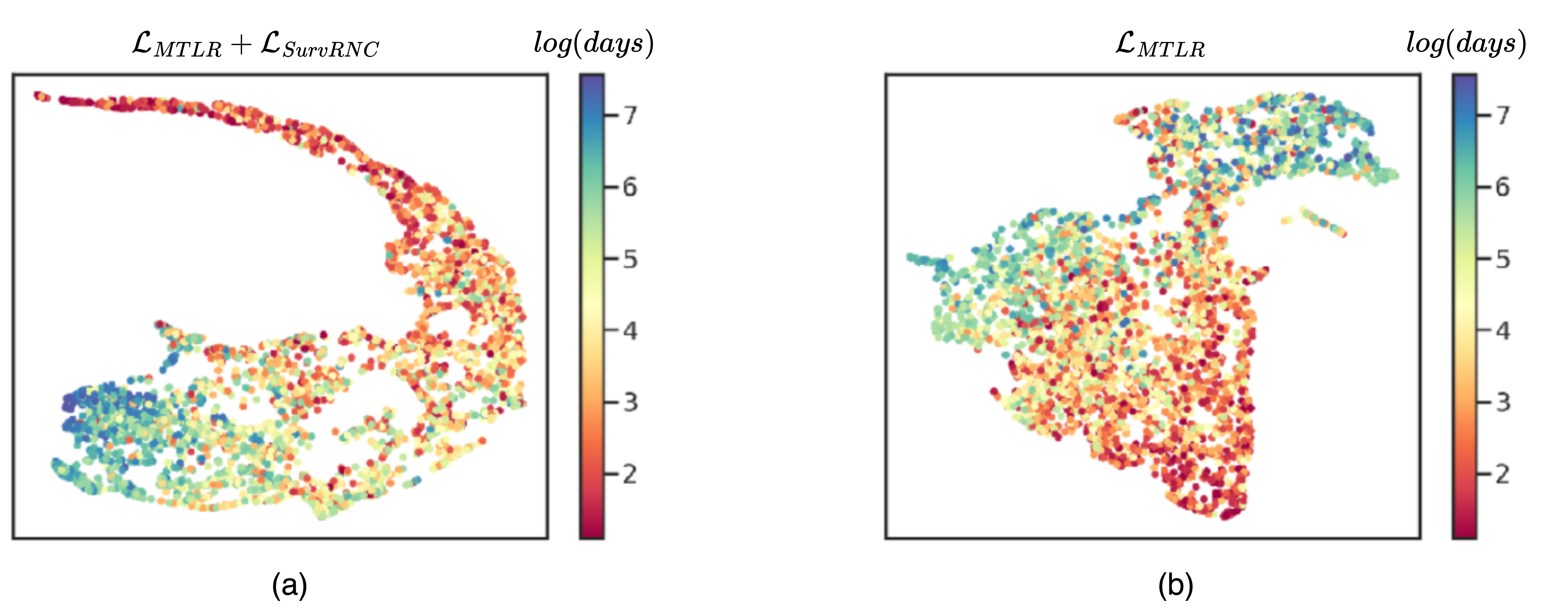}
    \caption{\textbf{UMAP Visualization:} $\mathcal{L}_{SurvRNC}$ with the native loss function of a DeepMTLR survival model on the SUPPORT dataset \cite{knaus1995support} shows a better continuous latent feature representation (a) as compared to using only the native loss function (b). }
    \label{fig:support}
\end{figure}

%% file: main.bbl
\begin{thebibliography}{10}
\providecommand{\url}[1]{\texttt{#1}}
\providecommand{\urlprefix}{URL }
\providecommand{\doi}[1]{https://doi.org/#1}

\bibitem{pahoWorldCancer}
{W}orld {C}ancer {D}ay 2023: {C}lose the care gap --- paho.org. \url{https://www.paho.org/en/campaigns/world-cancer-day-2023-close-care-gap}, [Accessed 25-02-2024]

\bibitem{andrearczyk2021overview}
Andrearczyk, V., Oreiller, V., Boughdad, S., Rest, C.C.L., Elhalawani, H., Jreige, M., Prior, J.O., Valli{\`e}res, M., Visvikis, D., Hatt, M., et~al.: Overview of the hecktor challenge at miccai 2021: automatic head and neck tumor segmentation and outcome prediction in pet/ct images. In: 3D head and neck tumor segmentation in PET/CT challenge, pp. 1--37. Springer (2021)

\bibitem{arefan2020deep}
Arefan, D., Mohamed, A.A., Berg, W.A., Zuley, M.L., Sumkin, J.H., Wu, S.: Deep learning modeling using normal mammograms for predicting breast cancer risk. Medical physics  \textbf{47}(1),  110--118 (2020)

\bibitem{buckley1979linear}
Buckley, J., James, I.: Linear regression with censored data. Biometrika  \textbf{66}(3),  429--436 (1979)

\bibitem{choi2022estimating}
Choi, S.R., Lee, M.: Estimating the prognosis of low-grade glioma with gene attention using multi-omics and multi-modal schemes. Biology  \textbf{11}(10), ~1462 (2022)

\bibitem{cox1972regression}
Cox, D.R.: Regression models and life-tables. Journal of the Royal Statistical Society: Series B (Methodological)  \textbf{34}(2),  187--202 (1972)

\bibitem{elmarakeby2021biologically}
Elmarakeby, H.A., Hwang, J., Arafeh, R., Crowdis, J., Gang, S., Liu, D., AlDubayan, S.H., Salari, K., Kregel, S., Richter, C., et~al.: Biologically informed deep neural network for prostate cancer discovery. Nature  \textbf{598}(7880),  348--352 (2021)

\bibitem{deepmtlr}
Fotso, S.: Deep neural networks for survival analysis based on a multi-task framework. ArXiv  \textbf{abs/1801.05512} (2018), \url{https://api.semanticscholar.org/CorpusID:13482950}

\bibitem{hu2022deep}
Hu, J., Yu, W., Dai, Y., Liu, C., Wang, Y., Wu, Q.: A deep neural network for gastric cancer prognosis prediction based on biological information pathways. Journal of Oncology  \textbf{2022} (2022)

\bibitem{huang2020deep}
Huang, Z., Johnson, T.S., Han, Z., Helm, B., Cao, S., Zhang, C., Salama, P., Rizkalla, M., Yu, C.Y., Cheng, J., et~al.: Deep learning-based cancer survival prognosis from rna-seq data: approaches and evaluations. BMC medical genomics  \textbf{13},  1--12 (2020)

\bibitem{ishwaran2008random}
Ishwaran, H., Kogalur, U.B., Blackstone, E.H., Lauer, M.S.: Random survival forests  (2008)

\bibitem{kim2019deep}
Kim, D.W., Lee, S., Kwon, S., Nam, W., Cha, I.H., Kim, H.J.: Deep learning-based survival prediction of oral cancer patients. Scientific reports  \textbf{9}(1), ~6994 (2019)

\bibitem{knaus1995support}
Knaus, W.A., Harrell, F.E., Lynn, J., Goldman, L., Phillips, R.S., Connors, A.F., Dawson, N.V., Fulkerson, W.J., Califf, R.M., Desbiens, N., et~al.: The support prognostic model: Objective estimates of survival for seriously ill hospitalized adults. Annals of internal medicine  \textbf{122}(3),  191--203 (1995)

\bibitem{deephit}
Lee, C., Zame, W., Yoon, J., Van~der Schaar, M.: {DeepHit}: A deep learning approach to survival analysis with competing risks. Proc. Conf. AAAI Artif. Intell.  \textbf{32}(1) (Apr 2018)

\bibitem{mariotto2014cancer}
Mariotto, A.B., Noone, A.M., Howlader, N., Cho, H., Keel, G.E., Garshell, J., Woloshin, S., Schwartz, L.M.: Cancer survival: an overview of measures, uses, and interpretation. Journal of the National Cancer Institute Monographs  \textbf{2014}(49),  145--186 (2014)

\bibitem{meng2023merging}
Meng, M., Bi, L., Fulham, M., Feng, D., Kim, J.: Merging-diverging hybrid transformer networks for survival prediction in head and neck cancer. In: International Conference on Medical Image Computing and Computer-Assisted Intervention. pp. 400--410. Springer (2023)

\bibitem{meng2022deepmts}
Meng, M., Gu, B., Bi, L., Song, S., Feng, D.D., Kim, J.: Deepmts: deep multi-task learning for survival prediction in patients with advanced nasopharyngeal carcinoma using pretreatment pet/ct. IEEE Journal of Biomedical and Health Informatics  \textbf{26}(9),  4497--4507 (2022)

\bibitem{ng2023deep}
Ng, C.W., Wong, K.K.: Deep learning can predict prognosis and endocrine therapy response in breast cancer patients from h\&e staining based on estrogen receptor signaling activity  (2023)

\bibitem{paszke2019pytorch}
Paszke, A., Gross, S., Massa, F., Lerer, A., Bradbury, J., Chanan, G., Killeen, T., Lin, Z., Gimelshein, N., Antiga, L., et~al.: Pytorch: An imperative style, high-performance deep learning library. Advances in neural information processing systems  \textbf{32} (2019)

\bibitem{polsterl2015fast}
P{\"o}lsterl, S., Navab, N., Katouzian, A.: Fast training of support vector machines for survival analysis. In: Machine Learning and Knowledge Discovery in Databases: European Conference, ECML PKDD 2015, Porto, Portugal, September 7-11, 2015, Proceedings, Part II 15. pp. 243--259. Springer (2015)

\bibitem{rebaud2022simplicity}
Rebaud, L., Escobar, T., Khalid, F., Girum, K., Buvat, I.: Simplicity is all you need: out-of-the-box nnunet followed by binary-weighted radiomic model for segmentation and outcome prediction in head and neck pet/ct. In: 3D Head and Neck Tumor Segmentation in PET/CT Challenge, pp. 121--134. Springer (2022)

\bibitem{saeed2021ensemble}
Saeed, N., Al~Majzoub, R., Sobirov, I., Yaqub, M.: An ensemble approach for patient prognosis of head and neck tumor using multimodal data. In: 3D Head and Neck Tumor Segmentation in PET/CT Challenge, pp. 278--286. Springer (2021)

\bibitem{saeed2022tmss}
Saeed, N., Sobirov, I., Al~Majzoub, R., Yaqub, M.: Tmss: An end-to-end transformer-based multimodal network for segmentation and survival prediction. In: International Conference on Medical Image Computing and Computer-Assisted Intervention. pp. 319--329. Springer (2022)

\bibitem{wang2019artificial}
Wang, S., Yang, D.M., Rong, R., Zhan, X., Fujimoto, J., Liu, H., Minna, J., Wistuba, I.I., Xie, Y., Xiao, G.: Artificial intelligence in lung cancer pathology image analysis. Cancers  \textbf{11}(11), ~1673 (2019)

\bibitem{wang2021gpdbn}
Wang, Z., Li, R., Wang, M., Li, A.: Gpdbn: deep bilinear network integrating both genomic data and pathological images for breast cancer prognosis prediction. Bioinformatics  \textbf{37}(18),  2963--2970 (2021)

\bibitem{yu2011learning}
Yu, C.N., Greiner, R., Lin, H.C., Baracos, V.: Learning patient-specific cancer survival distributions as a sequence of dependent regressors. Advances in neural information processing systems  \textbf{24} (2011)

\bibitem{zha2024rank}
Zha, K., Cao, P., Son, J., Yang, Y., Katabi, D.: Rank-n-contrast: Learning continuous representations for regression. Advances in Neural Information Processing Systems  \textbf{36} (2024)

\bibitem{zheng2022preoperative}
Zheng, Y., Wang, F., Zhang, W., Li, Y., Yang, B., Yang, X., Dong, T.: Preoperative ct-based deep learning model for predicting overall survival in patients with high-grade serous ovarian cancer. Frontiers in Oncology  \textbf{12},  986089 (2022)

\end{thebibliography}
